\documentclass[letterpaper, 10 pt, conference]{ieeeconf}  

\IEEEoverridecommandlockouts                              

\usepackage{amsmath,amsfonts}
\usepackage{graphics} 
\usepackage{epsfig} 
\usepackage{times} 
\usepackage{amsmath} 
\usepackage{amssymb}  
\usepackage{algorithmic}
\usepackage{algorithm}
\usepackage{array}
\usepackage{booktabs}
\usepackage{caption}
\usepackage{multirow}
\usepackage{listings}
\usepackage{wrapfig}
\usepackage{textcomp}
\usepackage{stfloats}
\usepackage{url}
\usepackage{soul}
\usepackage{verbatim}
\usepackage{graphicx}
\usepackage{orcidlink}
\usepackage{cite}
\let\labelindent\relax
\usepackage{enumitem}
\usepackage{marginnote}
\usepackage{amsmath}
\usepackage{amssymb}

\title{\LARGE \bf
Physics-Conditioned Grasping for Stable Tool Use
}

\author{Noah Trupin$^{\dagger}$ \quad Zixing Wang$^{\dagger}$ \quad Ahmed H. Qureshi \vspace{0.3em} \\ {Purdue University}
\vspace{0.3em} \\ \small{\texttt{\{ntrupin, wang5389, ahqureshi\}@purdue.edu}}
\thanks{$^{\dagger}$ denotes equal contribution.}%
}

\begin{document}

\maketitle
\thispagestyle{empty}
\pagestyle{empty}

\begin{abstract}
    Tool use often fails not because robots misidentify tools, but because grasps cannot withstand task-induced wrench. Existing vision-language manipulation systems ground tools and contact regions from language yet select grasps under quasi-static or geometry-only assumptions. During interaction, inertial impulse and lever-arm amplification generate wrist torque and tangential loads that trigger slip and rotation. We introduce inverse Tool-use Planning (iTuP), which selects grasps by minimizing predicted interaction wrench along a task-conditioned trajectory. From rigid-body mechanics, we derive torque, slip, and alignment penalties, and train a Stable Dynamic Grasp Network (SDG-Net) to approximate these trajectory-conditioned costs for real-time scoring. Across hammering, sweeping, knocking, and reaching in simulation and on hardware, SDG-Net suppresses induced torque up to 17.6\%, shifts grasps below empirically observed instability thresholds, and improves real-world success by 17.5\% over a compositional baseline. Improvements concentrate where wrench amplification dominates, showing that robot tool use requires wrench-aware grasp selection, not perception alone.
\end{abstract}

\section{Introduction}

Robotic tool use extends embodiment: a hammer amplifies impulse~\cite{qin2023robot}, a brush aggregates distributed contact~\cite{xieimprovisation}, and a stick transfers force beyond reach~\cite{Toussaint2018DifferentiablePA, xu2023creative}. Yet in practice, tool-use systems often fail in a simple way: they select the correct tool and plan the correct motion, but the grasp slips or twists under interaction forces. The failure is mechanical rather than semantic.

Vision-language models (VLMs) have significantly improved open-vocabulary grounding for manipulation. Robots can identify novel tools, localize contact regions, and infer interaction directions from natural language~\cite{li2021grounded, huang2024copa, huang2023voxposer}. However, grasp selection is typically performed using geometry-only metrics or quasi-static force-closure assumptions, evaluated independently of the intended motion. These criteria ignore how forces induced by the task trajectory propagate through the grasp. Existing tool-use systems choose grasps based on geometry, but stability is determined by interaction wrench.

Tool use is fundamentally a wrench-transfer problem. Forces applied at remote contact points induce wrist torque $\tau = r \times F$, whose magnitude and tangential projection determine grasp stability. Robust tool use therefore depends on both lever-arm geometry and alignment.

\begin{figure}[t!]
    \vspace{5pt}
    \includegraphics[width=\linewidth]{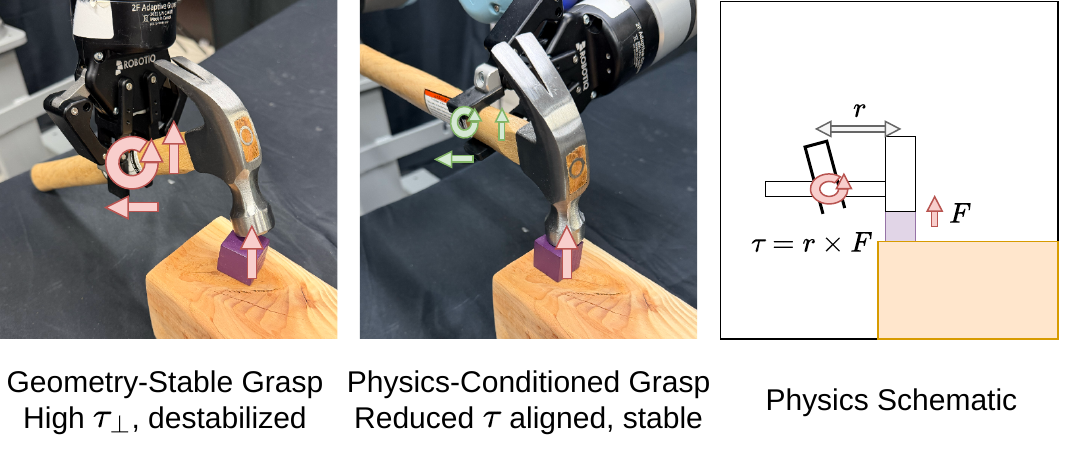}
    \caption{\textbf{Stable tool use as wrench transfer.} \textbf{Left:} Under identical contact location and hammer trajectory, a geometry-ranked grasp rotates about the gripper frame during peak impact due to lever-arm amplification. \textbf{Center:} The physics-conditioned grasp shortens the lever arm and aligns the jaw normal with the interaction normal, reducing predicted wrist torque below the instability threshold and preventing slip. \textbf{Right:} Schematic of $\tau = r \times F$ governing the observed rotation.}
    \label{fig:title}
    \vspace{-18pt}
\end{figure}

We address this gap by introducing inverse Tool-use Planning (iTuP), a framework that conditions grasp selection on predicted interaction wrench. Given grounded tool–object contact parameters and a short-horizon interaction trajectory, we analytically derive torque, slip, and alignment penalties from rigid-body mechanics. These penalties quantify how a candidate grasp amplifies or suppresses the induced wrench. We then train a Stable Dynamic Grasp Network (SDG-Net) to approximate these trajectory-conditioned costs from local geometry and motion features, enabling real-time evaluation over large candidate sets. We call this `inverse' planning because grasp selection is conditioned on the wrench implied by a specified interaction, rather than planning interaction around a fixed grasp.

Essentially, iTuP decouples semantic grounding from mechanical feasibility. VLMs determine what tool to use and where to interact; SDG-Net determines whether the grasp will survive the induced torque. Improvements in stability therefore arise from wrench-aware grasp scoring rather than changes in semantic reasoning or visual architecture.

We evaluate iTuP across impulse-driven (hammer, knock), multi-contact (sweep), and lever-arm-dominated (reach) regimes in simulation and hardware. SDG-Net reduces peak induced torque by up to 17.6\% and yields a 17.5\% improvement in real-world task success over a compositional VLM baseline. Ablations confirm that removing physics-conditioned grasp scoring reintroduces torque-driven failures even when grounding and trajectory planning remain unchanged.

Together, these results demonstrate that robust tool use benefits from explicit conditioning on predicted wrench transmission through the grasp. We formalize and validate this perspective through the following contributions: \begin{itemize}
    \item \textbf{Wrench-conditioned grasp formulation.} We formulate tool-use grasp selection as minimizing predicted interaction torque and slip induced by a task-conditioned trajectory.
    \item \textbf{Analytically derived penalties.} We derive physically grounded costs that scale with both impulse magnitude and lever-arm length.
    \item \textbf{Stable Dynamic Grasp Network (SDG-Net).} We introduce a learned surrogate that approximates trajectory-conditioned wrench costs for real-time evaluation.
    \item \textbf{Causal validation across regimes.} We show that reductions in predicted torque reduce slip and increase task success across impulse-driven and lever-arm-dominated interactions.
\end{itemize}

As highlighted by these contributions, iTuP provides a principled step towards robust, generalizable stable manipulation.

\section{Related Work}
\label{sec:relatedwork}

\subsection{Robot Tool Use}

Tool use extends robotic embodiment by enabling force transfer beyond direct grasping. Early approaches relied on structured affordance representations and symbolic reasoning~\cite{duan2025aha, huang2023voxposer, huang2024copa}, while later work incorporated differentiable physics, task-and-motion planning (TAMP), and learned dynamics models to improve generalization~\cite{pmlr-v78-rana17a, qin2021rapidly, fang2020learning, Toussaint2018DifferentiablePA}.

Despite these advances, most tool-use systems assume geometric grasp stability once force-closure is satisfied. Stability under impact, lever-arm amplification, and multi-contact aggregation is rarely treated as an explicit design constraint. We instead frame tool use as a wrench-transfer problem, where grasp feasibility must be evaluated under the forces induced by the task trajectory.

\subsection{Grasp Stability}

Classical grasp analysis evaluates stability using force-closure and wrench-space metrics under static equilibrium assumptions~\cite{mahler2016dex, mahler2017dex}. Learning-based methods such as Dex-Net, Contact-GraspNet, and related approaches predict robust 6-DoF grasps from point clouds~\cite{fang2020graspnet, wang2021graspness, sundermeyer2021contact, morrison2018closing, DBLP:journals/corr/PasGSP17, acronym2020, fang2023robust, morrison2019multiview, 1603.01564}, improving generalization in cluttered scenes.

However, nearly all grasp methods evaluate stability in isolation—typically for object pickup or quasi-static manipulation. Tool use differs fundamentally: forces are applied at a remote contact point, generating torque that scales with the lever arm between contact and grasp. A grasp stable for lifting may fail under interaction-induced wrench. Our formulation conditions grasp evaluation on predicted task-induced torque rather than geometry alone.

\subsection{Vision-Language Models in Manipulation}

Vision-language models enable open-vocabulary grounding, segmentation, and affordance reasoning for manipulation~\cite{kirillov2023segany, li2023semantic, Qi_2017_CVPR, li2022mask, pgvlm2024, zhang2024improvevisionlanguagemodel, yang2024guidinglonghorizontaskmotion, huang2023voxposer, wei2023chainofthoughtpromptingelicitsreasoning, yang2023setofmark, huang2024copa}. Recent systems integrate VLM reasoning with motion planning or end-to-end visuomotor control.

While effective at determining what tool to use and where to interact, these approaches typically delegate grasp selection to geometry-based metrics or learned policies that do not explicitly evaluate induced torque or slip under interaction. Our framework preserves semantic flexibility while introducing an explicit wrench-conditioned grasp evaluation layer.

\subsection{Physics-Aware Manipulation and Dynamic Interaction}

Physics-aware manipulation incorporates differentiable simulators, model-predictive control, or learned world models to predict object motion under contact~\cite{mittal2023orbit, Xiang_2020_SAPIEN, liang2024dreamitate, todorov2012mujoco, wang2025implicit}. These methods reason about interaction dynamics for trajectory planning or policy learning; however, grasp pose is typically fixed or selected using geometry-based metrics prior to dynamic reasoning. Conditioning grasp selection itself on predicted task-induced wrench remains underexplored.

We explicitly derive trajectory-conditioned torque and slip penalties from rigid-body mechanics and use them to score grasp candidates prior to execution.

\subsection{Summary of the Gap}

Prior work advances along largely independent axes: \begin{enumerate}[label=(\arabic*)]
    \item semantic grounding,
    \item geometric grasp stability, and
    \item physics-aware motion planning.
\end{enumerate}

The coupling between grasp selection and predicted interaction wrench remains underexplored. This coupling is essential for tool use, where lever-arm geometry and contact forces can amplify torque beyond grasp limits. Our work addresses this structural gap by conditioning grasp selection on predicted wrench transmission through the grasp.
    
\section{Methods}
\label{sec:methods}

\begin{figure}[t!]
    \centering
    \includegraphics[width=\linewidth]{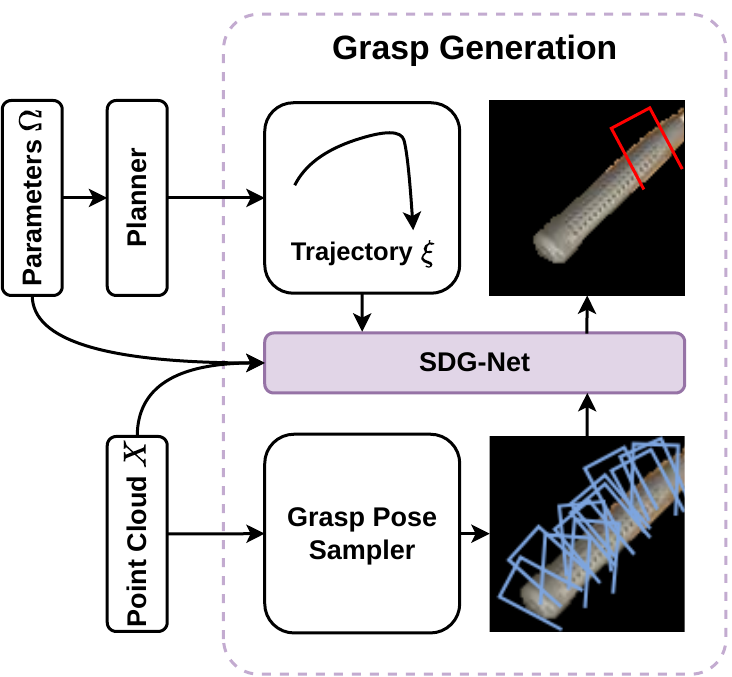}
    \caption{\textbf{Conditioned grasp generation.} Given fixed contact parameters and a short-horizon trajectory $\xi(t)$, candidate grasps are evaluated under predicted interaction wrench. SDG-Net approximates analytically derived torque, slip, and alignment penalties, enabling fast parallel scoring. Grasp selection minimizes predicted wrench amplification rather than geometric stability alone.}
    \label{fig:title}
    \vspace{-15pt}
\end{figure}

Tool use differs from object pickup in one fundamental respect: the grasp must transmit wrench generated at a remote contact point. When the end-effector executes a task-conditioned trajectory, the resulting impulse and follow-through motion produce forces and torques that are amplified by the lever arm between the tool contact and the grasp frame. A grasp that is stable for lifting may therefore fail under impact.

We formalize grasp selection as minimizing the predicted interaction wrench induced by a task trajectory. Our framework separates: \begin{enumerate}[label=(\arabic*)]
    \item \textbf{Semantic grounding} (what tool and contact to use)
    \item \textbf{Trajectory synthesis} (how the interaction unfolds)
    \item \textbf{Wrench-conditioned grasp evaluation} (whether the grasp survives the interaction).
\end{enumerate}

This section derives the trajectory-conditioned grasp cost, establishes its physical invariances, and introduces SDG-Net as a learned surrogate for real-time evaluation.

\subsection{Problem Formulation: Grasping Under Predicted Interaction Wrench}

We consider tool use as a \textit{wrench transfer problem}. A tool-use task is defined by: \begin{itemize}
    \item Scene observation $q \in \mathcal{Q}$,
    \item Object set $\mathcal{I}$ containing candidate tools and targets, and
    \item Natural language instruction $\phi \in \Phi$.
\end{itemize} The planner must produce \begin{enumerate}[label=(\arabic*)]
    \item A grasp $g \in SE(3)$ on a selected tool and
    \item A short-horizon end-effector trajectory $\xi(t),\,t \in [0, T]$, that executes the intended interaction.
\end{enumerate} Unlike conventional grasp planning, the grasp cannot be evaluated independently of the trajectory. The trajectory induces an interaction wrench $W_\mathrm{int}$ at the tool contact point, which is transmitted through the grasp to the wrist.

A grasp is successful if it (i) suppresses destabilizing torque at the wrist, (ii) minimizes tangential force projection at the gripper interface, and (iii) maintains alignment between gripper surface normals and the interaction normal during contact. We therefore define grasp selection as \begin{equation}
    g^* = \arg\min_{g \in \mathcal{G}} C(g \mid \xi, \Omega)
\end{equation} where $\mathcal{G}$ is a set of candidate grasps, $\Omega$ are task-conditioned contact parameters, and $C$ is a trajectory-conditioned wrench penalty. Unlike classical grasp planning, the cost $C(g \mid \xi, \Omega)$ is not intrinsic to the grasp geometry. It is induced by the interaction wrench generated when trajectory $\xi$ is executed. A grasp that is stable under quasi-static motion may become unstable under high-impulse follow-through. Thus, grasp quality is a function of predicted interaction dynamics, not geometry alone.

\subsection{Interaction Regimes}

We distinguish regimes by the relative magnitude of inertial forces and grasp reaction forces. With tool--object contact forces $F_\mathrm{ext}$ and internal grasp reactions $F_\mathrm{grasp}$, the \textit{quasi-static} regime satisfies $\|ma\| \ll \|F_\mathrm{grasp}\|$, the \textit{dynamic} regime satisfies $\|ma\| \gtrsim \|F_\mathrm{grasp}\|$, and the clustered regime aggregates multiple contacts via $\sum_i F_{\mathrm{ext}, i} = ma$.

Critically, instability is not determined solely by inertial dominance. Torque scales as $\|\tau\| = \|r \times F\|$, so even when $\|ma\| \ll \|F_\mathrm{grasp}\|$, large lever arms $\|r\|$ can amplify modest contact forces into destabilizing torque. Thus, grasp stability must be evaluated under predicted wrench transmission rather than static geometry alone.

\subsection{Trajectory-Conditioned Wrench Derivation}

Let the tool center of mass be $c_\mathrm{COM}$, tool--object contact point $c_\mathrm{obj}$, and lever arm $r = c_\mathrm{obj} - c_\mathrm{COM}$. For end-effector linear and angular velocities $(v, \omega)$, the relative contact velocity is \begin{equation*}
    v_c = v + \omega \times r.
\end{equation*}

Under rigid-body impact with restitution $e$, mass $m$, and inertia $I$, the normal impulse magnitude is \begin{equation*}
    J = -\frac{(1 + e)v_c \cdot n}{\frac{1}{m} + n \cdot \left[ I^{-1} (r \times n) \times r\right]}.
\end{equation*}

Exact inertial parameters are not directly measured in hardware. We estimate approximate mass and inertia from category priors inferred during object-level grounding, combined with geometric properties of the segmented mesh. Restitution and friction coefficients are assigned from category-dependent priors. These parameters need only be approximate, as SDG-Net learns to predict expected wrench amplification under modeling uncertainty.

The induced contact force over duration $\Delta t$ and resulting torque at the grasp frame are \begin{equation*}
    F = \frac{J}{\Delta t} n, \quad \tau = r \times F.
\end{equation*}

We use the normal impulse to parameterize the dominant torque amplification term. Decomposing the force relative to the gripper surface normal $n_\mathrm{finger}$,  \begin{equation*}
    \|\tau\| = \|r\| \cdot \|F_\bot\|, \quad F_t = F - (F \cdot n_\mathrm{finger}) n_\mathrm{finger},
\end{equation*} where $F_\bot$ is perpendicular to the lever arm and $F_t$ is tangential to the gripper interface. Stability therefore depends jointly on lever-arm length and tangential force projection, motivating the torque, slip, and alignment penalties defined in Sec.~\ref{sec:methods-cost}.

\subsection{Physics-Conditioned Grasp Cost}
\label{sec:methods-cost}

\begin{figure*}[t!]
    \vspace{5pt}
    \centering
    \scalebox{0.9}{
        \includegraphics[width=\textwidth]{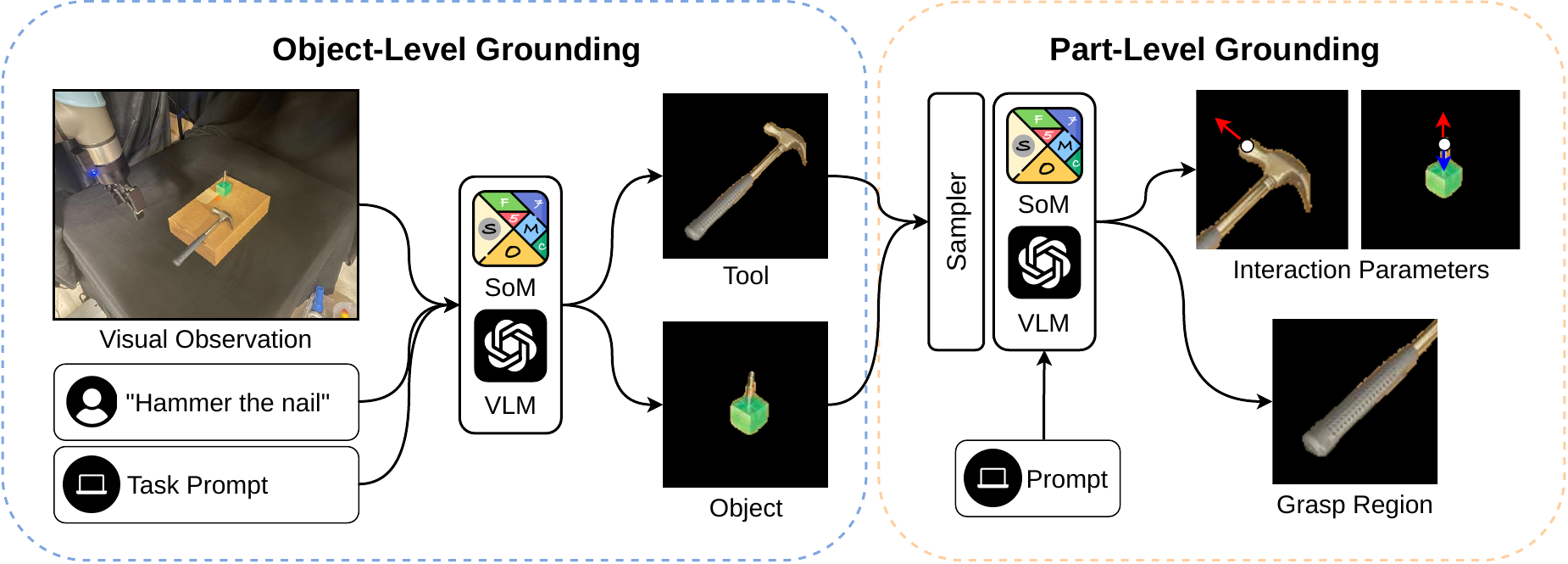}
    }
    \caption{\textbf{Two-level semantic grounding pipeline}. Object-level grounding identifies the relevant tool and target from natural language, while part-level grounding refines interaction geometry by selecting contact points and interaction direction. Grounding determines contact parameters $\Omega = \{c_\mathrm{tool}, c_\mathrm{obj}, n, d\}$ and a grasp region, but does not rank grasps. This separation ensures that stability improvements arise from physics-conditioned grasp evaluation rather than changes in semantic reasoning.}
    \label{fig:title}
    \vspace{-18pt}
\end{figure*}

We define three coupled penalties capturing distinct geometric failure mechanisms of wrench transmission: (i) alignment deviation between gripper and interaction normals, (ii) torque amplification due to lever-arm geometry, and (iii) slip onset due to tangential force projection.

\subsubsection{Interaction Torque}

Project torque onto wrist-sensitive axes: \begin{equation*}
    C_\tau(g) = \|\Pi_\bot \tau(g;\,\xi)\|_2
\end{equation*} where $\Pi_\bot$ projects onto axes orthogonal to jaw closure.

\subsubsection{Slip Penalty}

let $F_n, F_t$ be normal and tangential components of $F$. With friction coefficient $\mu$: \begin{equation*}
    C_s(g) = \max(0, \|F_t\| - \mu\|F_n\|)
\end{equation*}

\subsubsection{Alignment Deviation}

Let $n_\mathrm{finger}(g)$ denote gripper surface normal: \begin{equation*}
    C_\alpha(g) = \angle(n_\mathrm{finger}(g), n)
\end{equation*} where $\angle(\cdot, \cdot)$ denotes the angle between two vectors. This penalizes deviation between the gripper surface normal and the interaction normal, which increases tangential load and accelerates slip under repeated or impulsive contact.

These penalties yield a total cost \begin{equation}
    \label{eq:cost}
    C(g) = w_\tau C_\tau(g) + w_s C_s(g) + w_\alpha C_\alpha(g)
\end{equation} which satisfies the following physical properties: \begin{enumerate}[label=(C\arabic*)]
    \item Under rigid transformation, vectors $(v, \omega, r, n)$ transform covariantly while scalar norms remain invariant, preserving $C$.
    \item Noise robustness follows from local aggregation of geometric features and continuous dependence of $J$ on $v_c$.
    \item In the quasi-static limit $v, \omega \to 0$, the impulse vanishes and the cost reduces to alignment; in dynamic and clustered regimes, superposition of induced torques preserves consistency.
    \item The explicit torque and slip terms directly penalize destabilizing wrench components.
\end{enumerate}

\subsection{SDG-Net: Learning Trajectory-Conditioned Wrench Cost}

Although Eq.~\ref{eq:cost} defines a physically grounded objective, its exact evaluation is not directly observable in real-world settings: contact impulses, inertial parameters, and effective lever-arm geometry are only partially known prior to execution, and compliance introduces additional uncertainty. We therefore treat Eq.~\ref{eq:cost} as a structured objective and train SDG-Net to predict expected wrench amplification from local point-cloud features and trajectory-conditioned parameters.

Specifically, SDG-Net approximates \begin{equation*}
    \hat{C}(g \mid X, \xi, \Omega) \approx C(g),
\end{equation*} and is trained using \begin{equation*}
    \mathcal{L} = \sum_i \left[ (\hat{C}_{\tau, i} - C_{\tau, i})^2 + (\hat{C}_{s,i} - C_{s, i})^2 + (\hat{C}_{\alpha, i} - C_{\alpha, i})^2 \right].
\end{equation*}

At inference, grasp selection is performed via \begin{equation*}
    g^* = \arg\min_{g \in \mathcal{G}} \hat{C}(g)
\end{equation*}

SDG-Net thus estimates interaction-induced wrench under partial observability while preserving the physical structure of the cost.

\subsection{Separation of Semantics and Physics}

Our pipeline enforces a strict separation: \begin{itemize}
    \item VLM grounding selects tool and contact parameters,
    \item Trajectory synthesis defines interaction dynamics, and
    \item SDG-Net selects grasps under predicted wrench.
\end{itemize}

Grounding determines contact parameters and trajectory structure, but does not alter the physics-based cost definition.

\subsection{Full iTuP Pipeline}
\label{sec:methods-pipeline}

\begin{table*}[t!]
    \vspace{5pt}
    \centering
    \scalebox{1.25}{
        \begin{tabular}{lcccccc}\toprule
            \multirow{2}{*}{\textbf{Task}}&\multicolumn{5}{c}{\textbf{Methods}} \\ \cmidrule{2-6}
            & iTuP & w/o SDG-Net & w/o VLM & w/o 2-lvl gran. & CoPa \\
            \midrule
            Hammer nail & $\mathbf{50\%}$ & $30\%$ & $0\%$ & $10\%$ & $30\%$ \\
            Sweep toys  & $\mathbf{90\%}$ & $70\%$ & $10\%$ & $60\%$ & $70\%$ \\
            Knock tower & $\mathbf{90\%}$ & $70\%$ & $30\%$ & $80\%$ & $80\%$ \\
            Reach blocks & $\mathbf{80\%}$ & $70\%$ & $0\%$ & $30\%$ & $60\%$ \\
            \midrule
            \textbf{Total} & $\mathbf{77.5\%}$ & $60.0\%$ & $10.0\%$ & $45.0\%$ & $60.0\%$ \\
            \bottomrule
        \end{tabular}
    }
    \vspace{5pt}
    \caption{\textbf{Real-world task success across interaction regimes.} iTuP improves performance across tasks where torque amplification and slip dominate failure. Removing SDG-Net increases wrench-induced instability despite identical tool grounding and motion planning, isolating the contribution of physics-conditioned grasp scoring.}
    \label{tab:main}
    \vspace{-10pt}
\end{table*}

\begin{figure*}[b!]
    \vspace{-10pt}
    \includegraphics[width=\textwidth]{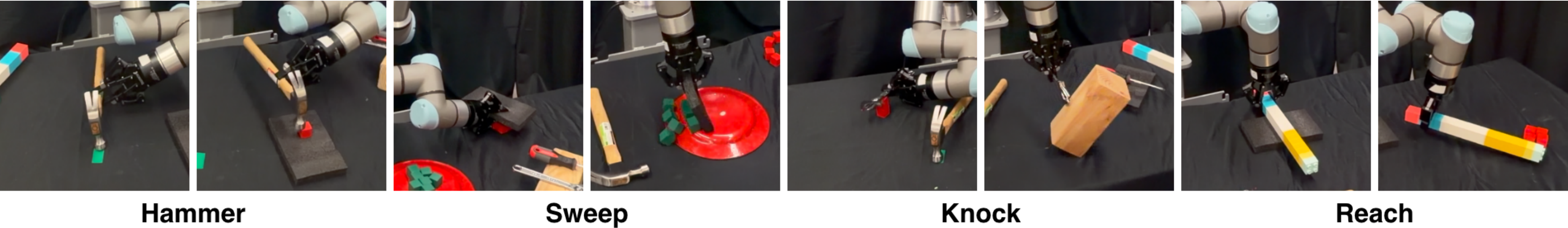}
    \caption{\textbf{Representative hardware tasks spanning distinct wrench-amplification regimes.} From left to right (grasp and interaction pairs): hammer (impulse-dominant impact), knock (impulse with lever-arm amplification), reach (lever-arm-dominated torque), and sweep (clustered multi-contact aggregation). These tasks isolate the methods of wrench amplification and enable regime-level evaluation of physics-conditioned grasp selection. See supplementary video for full execution.}
    \label{fig:experiments}
\end{figure*}

The full iTuP system operates in six stages: \begin{enumerate}[label=(\arabic*)]
    \item \textbf{Object-level grounding:} A VLM identifies tool and target objects from natural language and segmented and labeled scene observation~\cite{yang2023setofmark, zhang2022dino}.
    \item \textbf{Part-level grounding:} The VLM selects interaction-relevant contact points and direction, producing contact parameters $\Omega =\{c_\mathrm{tool}, c_\mathrm{obj}, n, d\}$.
    \item \textbf{Trajectory synthesis:} A short-horizon interaction trajectory $\xi(t)$ is generated from $\Omega$.
    \item \textbf{Grasp sampling:} A set $G \subset SE(3)$ of candidate grasps is produced.
    \item \textbf{Wrench-conditioned scoring:} SDG-Net predicts the trajectory-conditioned cost $\hat{C}(g \mid X, \xi, \Omega)$ for each $g \in G$.
    \item \textbf{Execution:} The grasp minimizing predicted cost is executed along $\xi(t)$.
\end{enumerate} In ablations isolating grasp modules, stages (1)--(3) are held fixed. Performance differences therefore isolate the contribution of physics-conditioned grasp scoring rather than semantic reasoning.

In summary, iTuP treats grasp selection as a function of induced interaction wrench. By conditioning stability on predicted dynamics rather than static geometry, the framework bridges semantic grounding and mechanical feasibility without conflating their roles.

\section{Experiments \& Results}
\label{sec:result}

\subsection{Experimental Overview}

We evaluate iTuP across simulation and hardware to answer: (1) does wrench-conditioned grasp selection reduce induced torque, (2) does torque reduction correlate with reduced slip and higher task success, and (3) are gains concentrated in torque-amplified regimes? We treat torque-induced instability as the primary failure mechanism.

In controlled experiments (Sec.~\ref{sec:result-simulation}, \ref{sec:result-isolation}), tool identity, contact location, and trajectory are held fixed; only grasp pose varies. This isolates the effect of wrench-conditioned grasp scoring from semantic grounding and trajectory generation to highlight the contribution of SDG-Net. In full-pipeline evaluation (Sec.~\ref{sec:result-real}), all modules operate jointly and success reflects both semantic correctness and mechanical feasibility.

We evaluate four representative tasks: hammer (impact), sweep (multi-contact), knock (impulse and lever arm), and reach (lever-arm-dominated extension) (Fig.~\ref{fig:experiments}). These tasks are selected to stress wrench amplification through different geometric mechanisms rather than semantic variation and span the regimes defined in Sec.~\ref{sec:methods}. In hardware, the regime failures manifest as visible rotation or slip during peak impact.

We emphasize that torque and slip measurements in Sec.~\ref{sec:result-simulation} isolate grasp scoring under fixed trajectories, while success rates in Sec.~\ref{sec:result-real} reflect the full decision pipeline.

\subsection{Simulation: Wrench Reduction Under Controlled Dynamics}
\label{sec:result-simulation}

We first measure induced wrench statistics in simulation using Isaac Sim~\cite{mittal2023orbit}. For each trial, we record peak induced wrist torque $\tau_\mathrm{max}$, slip magnitude $s_\mathrm{max}$, and tool-axis deviation $\alpha_\mathrm{max}$. Torque is measured at the wrist frame, slip as relative displacement between tool and gripper, and alignment deviation measures angular drift from intended interaction axis. Slip is computed from the relative transform between the tool mesh and gripper frame in simulation; although small in magnitude, increases in $s_\mathrm{max}$ are predictive of threshold crossing into visible tool drift in hardware.

We compare against GQ-CNN (Dex-Net 2.0)~\cite{mahler2016dex, mahler2017dex} (force closure scoring) and GraspNet~\cite{fang2020graspnet, fang2023anygrasp} (learned geometric grasp ranking). All baselines use identical tool--object contacts and trajectories.

\subsubsection{Torque Reduction}

Across tasks, SDG-Net reduces peak induced torque by up to \textbf{17.6\%} relative to geometry-based baselines. SDG-Net reduces the lever-arm-induced geometric torque amplification, explaining the substantial improvement observed in reach despite quasi-static motion profiles. As shown in Tab.~\ref{tab:sdgnet}, peak torque in Hammer drops from 8.01 Nm (GQ-CNN) to 6.60 Nm (SDG-Net), while corresponding slip decreases proportionally.

\subsubsection{Torque--Failure Correlation}

Fig.~\ref{fig:torque}-A visualizes the torque--slip phase across tasks with identical points and trajectories. Slip magnitude increases approximately monotonically with peak induced wrist torque, and failure cases ($\times$) cluster in the high-torque region, consistent with (C2). As such, torque amplification precedes observable slip and alignment drift, validating the wrench-transfer hypothesis.

As shown in Fig.~\ref{fig:torque}-B, failure probability exhibits a sharp transition near $\tau_\mathrm{max} \approx 6.9 \text{ Nm}$ across simulation trials. This narrow band corresponds with the friction-limited instability boundary predicted in Sec.~\ref{sec:methods-cost}. Above this region, slip and alignment drift increase rapidly, coinciding with (C4). Because grasp instability is thresholded by friction limits, modest torque suppression can shift grasps across this boundary, producing disproportionate gains in success. iTuP reduces occupancy of the high-instability region rather than incrementally improving stability within it.

\subsection{Real-World Evaluation: Regime-Specific Success}
\label{sec:result-real}

\begin{table*}[t!]
    \vspace{5pt}
    \centering
    \scalebox{0.92}{
        \begin{tabular}{l*{3}{ccc}}\toprule
            \multirow{2}{*}{\textbf{Task}} 
            & \multicolumn{3}{c}{\textbf{SDG-Net}} 
            & \multicolumn{3}{c}{\textbf{GQ-CNN}} 
            & \multicolumn{3}{c}{\textbf{GraspNet}} \\ 
            \cmidrule(lr){2-4}\cmidrule(lr){5-7}\cmidrule(lr){8-10}
            & $\tau \downarrow$ & $s \downarrow$ & $\alpha \downarrow$
            & $\tau \downarrow$ & $s \downarrow$ & $\alpha \downarrow$
            & $\tau \downarrow$ & $s \downarrow$ & $\alpha \downarrow$ \\
            \midrule
            Hammer & $\mathbf{6.60 \pm 0.34}$ & $\mathbf{0.01 \pm 0.00}$ & $\mathbf{0.15 \pm 0.02}$
                   & $8.01 \pm 0.28$          & $0.03 \pm 0.01$          & $0.23 \pm 0.03$
                   & $7.35 \pm 0.25$          & $0.03 \pm 0.00$           & $0.20 \pm 0.01$ \\
            Sweep  & $\mathbf{4.30 \pm 0.52}$ & $\mathbf{0.00 \pm 0.00}$ & $\mathbf{0.08 \pm 0.00}$
                   & $5.21 \pm 0.23$          & $0.03 \pm 0.00$          & $0.12 \pm 0.01$
                   & $\mathbf{5.02 \pm 0.20}$ & $0.02 \pm 0.00$          & $0.10 \pm 0.00$ \\
            Knock  & $\mathbf{5.24 \pm 0.66}$ & $\mathbf{0.01 \pm 0.00}$ & $\mathbf{0.12 \pm 0.00}$
                   & $\mathbf{5.93 \pm 0.75}$ & $0.05 \pm 0.01$          & $0.15 \pm 0.01$
                   & $\mathbf{5.85 \pm 0.52}$ & $0.04 \pm 0.00$          & $\mathbf{0.12 \pm 0.00}$ \\
            Reach  & $\mathbf{6.04 \pm 0.42}$ & $\mathbf{0.03 \pm 0.01}$ & $\mathbf{0.13 \pm 0.03}$
                   & $7.63 \pm 0.65$          & $0.05 \pm 0.00$          & $0.20 \pm 0.01$
                   & $6.98 \pm 0.30$          & $\mathbf{0.05 \pm 0.01}$ & $\mathbf{0.15 \pm 0.00}$ \\
            \bottomrule
        \end{tabular}
    }
    \caption{\textbf{Simulation measurements of peak wrist torque $\tau$, slip magnitude $s$, and alignment drift $\alpha$ under identical contact points and trajectories.} SDG-Net consistently reduces induced torque relative to geometry-based baselines. Increased torque correlates with increased slip and alignment drift, empirically validating the torque–failure causal chain.}
    \label{tab:sdgnet}
    \vspace{-10pt}
\end{table*}

\begin{figure}[b!]
    \vspace{-15pt}
    \includegraphics[width=\linewidth]{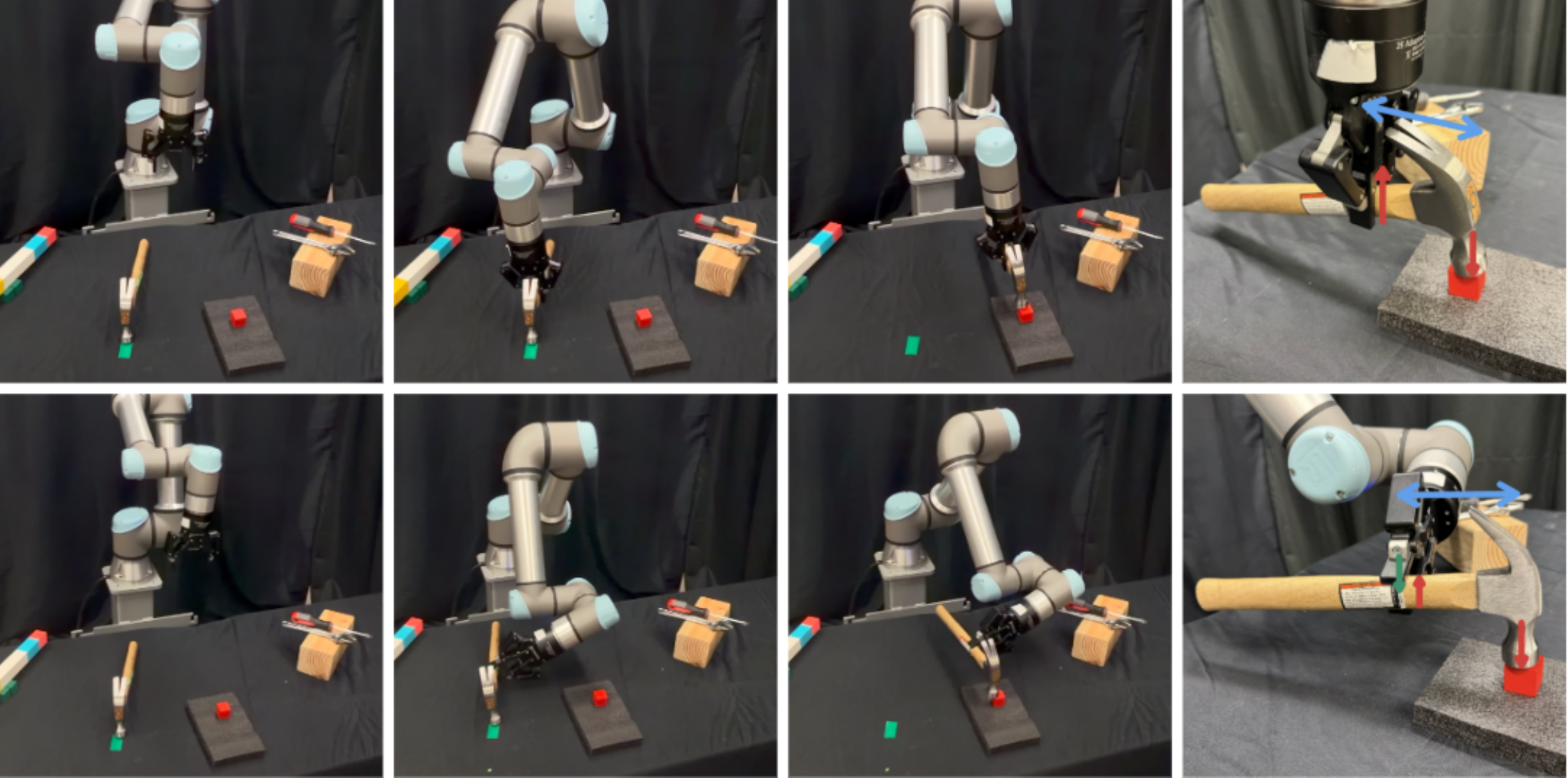}
    \caption{\textbf{Geometry-ranked vs. physics-conditioned grasps under identical hammer interactions.} Baseline grasps (top) amplify torque via misalignment and longer lever arm. Physics-conditioned grasps (bottom) align the interaction normal and shorten $r$, reducing induced wrench and preventing rotation. Final panels annotate $F$ (red, at head), $r$ (blue), and induced $\tau_\bot$ (red, at grasp) with suppression (green). See supplementary video for full execution.}
    \label{fig:hammer}
\end{figure}


We evaluate iTuP and ablations on a UR5e + Robotiq 2F-85 platform with $n = 10$ trials per task and report exact binomial proportions. We compare (i) \textbf{w/o SDG-Net}, replacing wrench-conditioned scoring with static force-closure; (ii) \textbf{w/o VLM}, removing semantic grounding; (iii)  \textbf{w/o 2-level gran.}, removing hierarchical refinement; and (iv) \textbf{CoPa}, a compositional VLM baseline. Success reflects joint semantic and mechanical performance across the entire decision pipeline, while ablations isolate module-level contributions. We further corroborate hardware trends using simulation-measured wrench statistics (Tab.~\ref{tab:sdgnet}), which exhibit consistent torque--slip scaling, and analyze downstream effects of torque reduction (Fig.~\ref{fig:torque}).


\subsubsection{Regime-Split Success Rates}

In reach, despite low acceleration, the extended lever arm produces substantial geometric torque amplification; SDG-Net mitigates this effect, improving stability even in a nominally quasi-static task. In the dynamic tasks (hammer, knock), removing SDG-Net increases torque-induced failures: in hammer, success drops from 50\% to 30\% without SDG-Net (Tab.~\ref{tab:main}), and failures manifest in hardware as visible tool rotation or gripper--tool drift at peak impact and follow-through. Fig.~\ref{fig:hammer} shows the mechanism qualitatively under the same interaction: geometry-ranked grasps amplify torque through misalignment and longer effective lever arm $r$, while physics-conditioned grasps align the jaw with the interaction normal and shorten $r$, suppressing induced $\tau_\bot$; this matches the higher peak induced torques observed for non-SDG grasps in simulation (Tab.~\ref{tab:sdgnet}). Finally, in the clustered-contact sweep task, cumulative contacts raise tangential loading and slip-driven failures increase with the removal of SDG-Net. Consistent with this regime analysis, Fig.~\ref{fig:torque}-C summarizes that tasks with larger torque suppression (e.g. hammer through $\|F\|$ and reach through $\|r\|$) exhibit larger failure-rate reductions relative to CoPa, indicating that the improvements track wrench suppression rather than task semantics.

\subsubsection{Structural Contribution of SDG-Net}

As shown in Tab.~\ref{tab:main}, overall success improves by 17.5\% over CoPa. Importantly, CoPa often selects the correct tool; failures arise from torque-induced grasp rotation or slip during impact. The improvement therefore reflects reduction of a specific mechanical failure mode rather than improved perception.

\subsection{Isolation from VLM Effects}
\label{sec:result-isolation}

To isolate grasp scoring from semantic improvements, we fix tool identity, contact points, interaction direction, and trajectory profile. Under this controlled comparison, only grasp pose varies. As reflected in Tab.~\ref{tab:sdgnet}, SDG-Net grasps exhibit lower torque amplification, while static grasps fail during high-impulse phases. This confirms that improvements arise from physics-conditioned grasp selection, not visual architecture changes.

\subsection{Robustness in Cluttered Scenes}

\begin{figure*}
    \vspace{5pt}
    \centering
    \includegraphics[width=\textwidth]{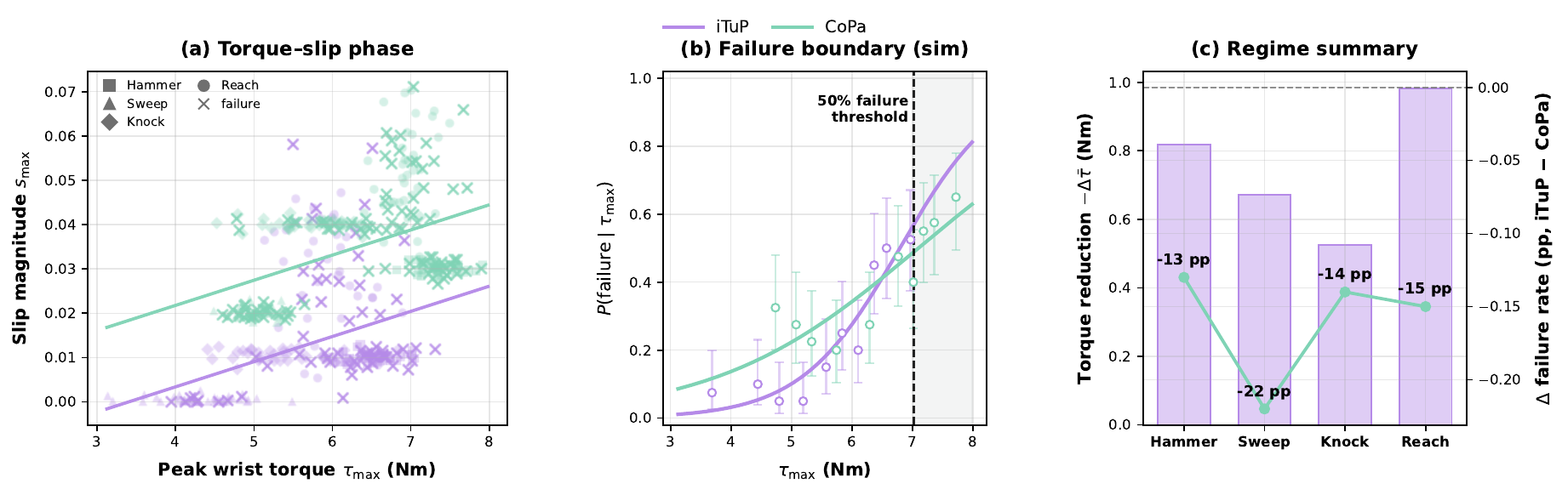}
    \caption{\textbf{Torque-failure phase diagram validating wrench-transfer causality.} \textbf{(a)} Under identical contact points and trajectories, slip increases approximately monotonically with peak induced wrist torque $\tau_\mathrm{max}$, and failures ($\times$) concentrate at higher torque. iTuP shifts the grasp distribution toward lower-torque regions. \textbf{(b)} Simulation failure probability rises sharply beyond a narrow torque band; the dashed line marks the empirical 50\% failure threshold. \textbf{(c)} Across interaction regimes, torque suppression predicts reductions in failure rate (percentage points, iTuP $-$ CoPa), indicating that wrench-conditioned grasp scoring removes a dominant mechanical failure mode.}
    \label{fig:torque}
    \vspace{-12pt}
\end{figure*}

\begin{wraptable}{r}{4.5cm}
\vspace{-10pt}
\centering
\begin{tabular}{lccccccc}\toprule
            \multirow{2}{*}{\textbf{Task}}&\multicolumn{2}{c}{\textbf{Architecture}}\\ \cmidrule{2-3}
            &\multicolumn{1}{c}{iTuP}&\multicolumn{1}{c}{CoPa} \\
            \midrule
            Hammer & $\mathbf{40\%}$ & $20\%$ \\
            Sweep & $\mathbf{90\%}$ & $70\%$ \\
            Knock & $\mathbf{90\%}$ & $70\%$ \\
            Reach & $\mathbf{60\%}$ & $30\%$ \\
            \midrule
            \textbf{Total} & $\mathbf{70.0\%}$ & $47.5\%$ \\
            \bottomrule
\end{tabular}
\caption{\textbf{Full-pipeline performance in cluttered scenes.} VLM baselines often select the correct tool, but static grasp metrics amplify torque through longer lever arms and lead to grasp rotation and slip. Physics-conditioned grasp selection yields substantial improvements.} \label{tab:clutter}
\vspace{-15pt}
\end{wraptable}

We further evaluate cluttered multi-tool scenes. In clutter, CoPa often selects correct tools but grasps suboptimal surfaces, static grasp metrics fail when lever-arm length increases, and SDG-Net selects handle-biased grasps that shorten the effective lever arm $r$, directly reducing induced torque. Success improves by over \textbf{22\%} compared to CoPa. The largest improvements occur in hammering and reaching, where torque amplification is dominant, and persist across varying scene layouts, consistent with (C1).

\subsection{Summary of Experimental Findings}

Overall, our experiments demonstrate that explicit conditioning on both lever-arm torque amplification and alignment-dependent force projection produces regime-dependent gains. In simulation, SDG-Net reduces induced torque by up to 17.6\%, directly shifting grasps away from the slip boundary. In hardware, this translates to a 17.5\% improvement in task success over CoPa and a 17.5\% improvement over the static-grasp ablation. Notably, improvements concentrate in regimes where torque amplification and misalignment-induced tangential loading dominate, either through inertial impulse (hammer, knock) or long lever arms (reach), consistent with the theoretical role of trajectory-conditioned torque suppression. In low-wrench interactions where $\|\tau\|$ remains small, SDG-Net performs comparably to static grasp metrics, consistent with the predicted quasi-static limit of the cost function and property (C3), where the cost reduces to alignment in the quasi-static limit.

Fig.~\ref{fig:torque} unifies the pattern: torque scales slip (Fig.~\ref{fig:torque}-A), torque crossing predicts failure (Fig.~\ref{fig:torque}-B), and torque suppression predicts regime-level success gains (Fig.~\ref{fig:torque}-C). These results indicate that SDG-Net reduces a dominant wrench-transfer failure mode.


\section{Conclusion \& Limitations}
\label{sec:conclusion}

Tool-use failures often arise not from semantic misidentification, but from mechanical instability under task-induced wrench. A robot may select the correct tool and motion yet still fail because the chosen grasp amplifies torque or projects force tangentially across the contact interface. We formalize this as a wrench-transfer problem: grasp stability depends on how interaction forces propagate through the grasp under the intended trajectory.

We introduced inverse Tool-use Planning (iTuP), which conditions grasp selection on predicted interaction wrench. By deriving torque, slip, and alignment penalties from rigid-body mechanics and approximating them with SDG-Net, we evaluate candidate grasps under the dynamics they intend to execute. This separates semantic grounding from mechanical feasibility while preserving real-time evaluation.

Across impulse-driven, multi-contact, and lever-arm-dominated regimes, higher induced torque correlates with increased slip and task failure. SDG-Net suppresses torque amplification and shifts grasps away from empirical instability thresholds, yielding consistent real-world gains concentrated in wrench-amplified regimes and comparable performance in the quasi-static limit.

Our model captures rigid-body impact and frictional slip but does not explicitly model compliance or long-horizon trajectory optimization. Grounding errors propagate into wrench estimation. Future work may integrate compliance-aware wrench estimation and joint optimization of grasp and trajectory under uncertainty.





\bibliographystyle{IEEEtran}
\bibliography{reference}

\end{document}